\documentclass[letterpaper,twocolumn,10pt]{article}
\usepackage{usenix}
\usepackage{tikz}
\usepackage{amsmath}
\usepackage{algorithm}
\usepackage{algpseudocode}
\usepackage{svg}
\usepackage{pgfplots}
\usepackage{booktabs}
\usepackage{multirow}
\usepackage{hyperref}
\usepackage{tabularx}
\usepackage{subfigure}
\usepackage{float}
\usepackage{makecell}
\usepackage{tikz-cd}
\usepackage{pgfplots}
\usepackage{enumitem}
\usepackage{adjustbox}
\usepackage{balance}
\usepackage{subcaption}
\usepackage{amsfonts}
\usepackage{natbib}
\pgfplotsset{compat=1.18}

\usepackage{fancyhdr}
\fancypagestyle{firstpage}{
  \fancyhf{} 
  \fancyhead[L]{Appeared at MLSys 2025 Young Professionals Symposium}
   
}

\makeatletter
\renewcommand{\paragraph}{%
  \@startsection{paragraph}{4}%
  {\z@}{1.25ex \@plus 1ex \@minus .2ex}{-1em}%
  {\normalfont\normalsize\bfseries}%
}
\makeatother

\newcommand{\bella}[0]{\textsc{Bella}}

\begin{document}

\date{\vspace{-4ex}}

\title{\vspace*{-0.5in} Trust by Design: Skill Profiles for Transparent, Cost-Aware LLM Routing}

\author{
{\rm Mika Okamoto\thanks{Corresponding author email: \href{mokamoto7@gatech.edu}{mokamoto7@gatech.edu}}, \quad Ansel Kaplan Erol, \quad Glenn Matlin}\\
\vspace{2 ex}
\textit{Georgia Institute of Technology}
\vspace{-2em} 
}
\maketitle
\thispagestyle{firstpage}

\vspace{-9ex}
\maketitle
\thispagestyle{firstpage}

\begin{abstract}
How should large language model (LLM) practitioners select the right model for a task without wasting money? We introduce \bella~(Budget-Efficient LLM Selection via Automated skill-profiling), a framework that recommends optimal LLM selection for tasks through interpretable skill-based model selection. Standard benchmarks report aggregate metrics that obscure which specific capabilities a task requires and whether a cheaper model could suffice. \bella~ addresses this gap through three stages: (1) decomposing LLM outputs and extract the granular skills required by using critic-based profiling, (2) clustering skills into structured capability matrices, and (3) multi-objective optimization to select the right models to maximize performance while respecting budget constraints. \bella~provides natural-language rationale for recommendations, providing transparency that current black-box routing systems lack. We describe the framework architecture, situate it within the landscape of LLM routing and evaluation, and discuss its application to financial reasoning as a representative domain exhibiting diverse skill requirements and cost-variation across models. Our framework enables practitioners to make principled and cost-performance trade-offs for deploying LLMs.

\end{abstract}

\section{Introduction}

The rapid expansion of Large Language Models (LLM) has created a critical deployment consideration for practitioners: \textit{how can practitioners select, and trust that they selected, the most suitable LLM for a task under real-world resource constraints?}
Selecting the right model requires understanding both task-specific capabilities and cost constraints, a challenge that current evaluation paradigms fail to address.

Current evaluation paradigms are insufficient for building human trust with these deployment decisions. Standard benchmarks report aggregate metrics like accuracy or F1 scores, but obscure which skills are needed to succeed on a specific task. Recent fine-grained evaluation methods~\citep{evaltree2025, moayeri2024skill} provide capability insights but ignore deployment constraints. Model routing frameworks~\citep{graphrouter2025, ong2024routellm} optimize for cost but operate as black boxes without interpretable rationale. Without comprehensive understanding of model selection, practitioners over- or under-provision resources, leading to wasted expenditure or suboptimal performance.

\begin{figure}[t]
\includegraphics[width=\columnwidth]{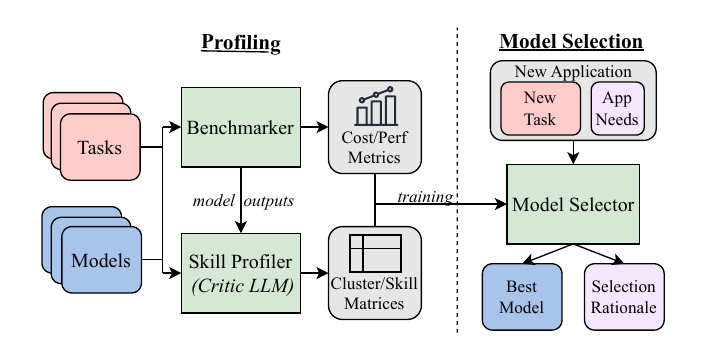}
\caption{\bella's architecture. A critic model analyzes LLM outputs on benchmark tasks to extract skill profiles. Given a new task and budget constraint, \bella~transparently selects the cost-optimal choice.}
\label{fig:architecture}
\vspace{-1 ex}
\end{figure}

We introduce \bella~(Budget-Efficient LLM Selection via Automated skill-profiling), a framework that addresses these gaps. Our contributions are:

\begin{itemize}[itemsep=2pt, leftmargin=*]
\item \textbf{Critic-based skill profiling:} A method for extracting interpretable, multi-skill capability profiles from LLM outputs without relying on pre-defined taxonomies.
\item \textbf{Constraint-aware selection:} An optimization algorithm that maximizes model performance within user-defined budget limits and provides a clear, interpretable rationale for the chosen model.
\item \textbf{Evaluation framework:} A leave-one-out method for assessing skill-based model selection using trusted, pre-existing benchmarks.
\end{itemize}

\noindent Although demonstrated in financial reasoning, the architecture of \bella~ generalizes to any domain with cost-performance trade-offs between any desired models. In doing so, \bella~offers a principled path toward transparent, resource-aware model deployment in real-world settings.

\newpage
\section{Background and Problem Setting}

\subsection{The Model Selection Challenge}

Consider a financial analysis application that requires numerical reasoning over tabular data, temporal understanding of market events, and verification of factual claims. Teams face a decision: use GPT-5 for maximum accuracy at \$10 per million tokens, or could Llama-3.3-70B at \$0.88 per million tokens\footnote{API pricing is obtained from TogetherAI.}---11$\times$ cheaper---achieve acceptable performance?

\begin{figure}[t]
\centering
\includegraphics[width=\columnwidth]{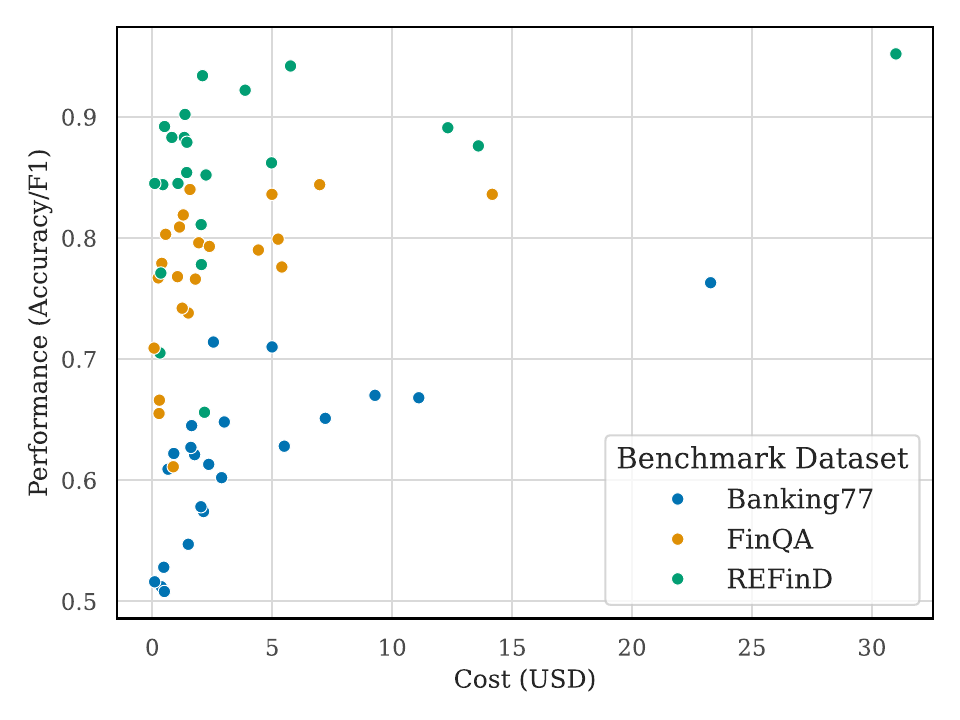}
\caption[capt for f1]{Cost-performance trade-offs across LLMs on financial reasoning benchmarks. More expensive models achieve higher average scores, but marginal gains diminish and capability profiles vary substantially at similar price points.\footnotemark}  
\label{fig:cost_performance}
\vspace{-1.5 ex}
\end{figure}

Aggregate benchmark scores cannot answer this question. A model achieving 85\% overall accuracy might excel at numerical reasoning but struggle with temporal logic, making it suitable for some financial tasks but problematic for others. Figure~\ref{fig:cost_performance} illustrates this challenge: while more expensive models achieve higher average performance, the marginal improvement per dollar diminishes rapidly, and substantial capability variation exists at each price point.

\footnotetext{Data for Figure \ref{fig:cost_performance} is sourced from FLaME~\citep{matlin-etal-2025-financial}.}

The fundamental problem is one of \textit{interpretable capability matching}: connecting what a task requires to what models can deliver, under constraints that real deployments impose.

\subsection{Desiderata for Model Selection}

An effective model selection framework should satisfy several properties:

\begin{enumerate}[itemsep=1pt, topsep=2pt,leftmargin=*]
\item \textbf{Skill-level granularity.} Selection should be based on specific capabilities (numerical reasoning, logical inference, factual recall) rather than aggregate scores that obscure which abilities matter for a given task.
\item \textbf{Cost awareness.} The framework should respect budget constraints explicitly, recommending models that maximize utility within specified cost or latency limits.
\item \textbf{Interpretability.} Practitioners should understand \textit{why} a model was recommended---which capabilities it possesses, which constraints it satisfies, and what trade-offs are involved. This transparency supports debugging, trust-building, and iterative refinement.
\item \textbf{Generalization.} The framework should transfer to new tasks without requiring extensive retraining. \bella's skill-based representations offer a path to this: if skills generalize across tasks, learned profiles can inform model selection on novel domains.
\end{enumerate}
\vspace{-3.2 ex}

\subsection{Related Work}

\begin{table}[t]
\centering
\small
\caption{Comparison of LLM selection and evaluation approaches across key dimensions.}
\label{tab:comparison}
\begin{tabular}{@{}lcccc@{}}
\toprule
\textbf{Approach} & \textbf{Skill} & \textbf{Cost-} & \textbf{Interp.} & \textbf{Multi-} \\
 & \textbf{Based} & \textbf{Aware} & & \textbf{Skill} \\
\midrule
FrugalGPT & \texttimes & \checkmark & \texttimes & \texttimes \\
RouteLLM & \texttimes & \checkmark & \texttimes & \texttimes \\
OptLLM & \texttimes & \checkmark & \texttimes & \texttimes \\
Hybrid-LLM & \texttimes & \checkmark & \texttimes & \texttimes \\
\midrule
FLASK & \checkmark & \texttimes & \checkmark & \checkmark \\
EvalTree & \checkmark & \texttimes & \checkmark & \texttimes \\
Skill-Slices & \checkmark & \texttimes & \checkmark & \checkmark \\
\midrule
\bella & \checkmark & \checkmark & \checkmark & \checkmark \\
\bottomrule
\end{tabular}
\end{table}

\paragraph{Fine-grained LLM evaluation}
Traditional benchmarks like MMLU and HumanEval~\citep{hendrycks2021mmlu, chen2021humaneval} report aggregate metrics but provide limited capability insight. Recent work addresses this gap: EvalTree~\citep{evaltree2025} builds hierarchical skill taxonomies but assumes one skill per instance, limiting applicability to multi-skill tasks. Skill-Slices~\citep{moayeri2024skill} clusters model rationales to reveal capability trade-offs but requires detailed chain-of-thought traces. CheckList~\citep{ribeiro2020checklist} and QualEval~\citep{murahari2024qualeval} provide behavioral testing and LLM-based critiques but lack per-instance skill attribution. In contrast, \bella~extracts multi-skill profiles at the instance level without requiring predefined taxonomies.

\paragraph{Model Routing and Selection} Routing frameworks dynamically select models to balance cost and quality. GraphRouter~\citep{feng2025graphrouter} learns routing policies via graph neural networks; RouteLLM~\citep{ong2024routellm} trains routers from preference data; FrugalGPT~\citep{chen2024frugalgpt} cascades through increasingly expensive models until confidence thresholds are met. These approaches improve efficiency but operate as black boxes without interpretable rationale for selection decisions. \bella~provides transparent, skill-based selection that explains why a model is recommended and generalizes to new tasks without extensive policy training.

\paragraph{Positioning of \bella.} Table~\ref{tab:comparison} situates \bella~relative to existing approaches across four dimensions: whether the method reasons about skills, incorporates cost constraints, provides interpretable rationale, and handles instances requiring multiple capabilities. \bella~uniquely combines skill-based profiling with cost-aware selection while maintaining interpretability. This makes \bella~ the first framework that enables practitioners to understand both \textit{why} a model is recommended and \textit{how} it satisfies their constraints.

\vspace{-2 ex}
\section{The \bella~Framework}

\begin{figure}[t]
\centering
\includegraphics[width=\columnwidth]{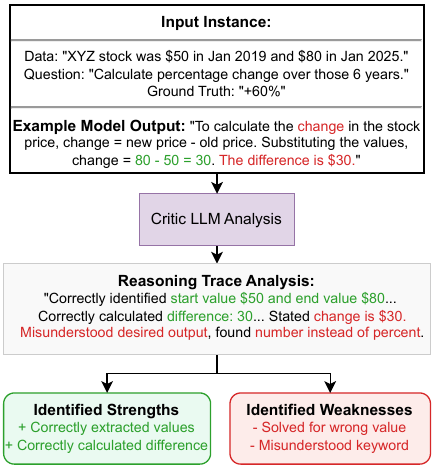}
\caption{Skill extraction via automated critique. Given a task, reference solution, and model output, the critic identifies demonstrated and missing skills.}
\label{fig:profiling_example}
\end{figure}

\begin{figure}[t]
\centering
\includegraphics[width=\columnwidth]{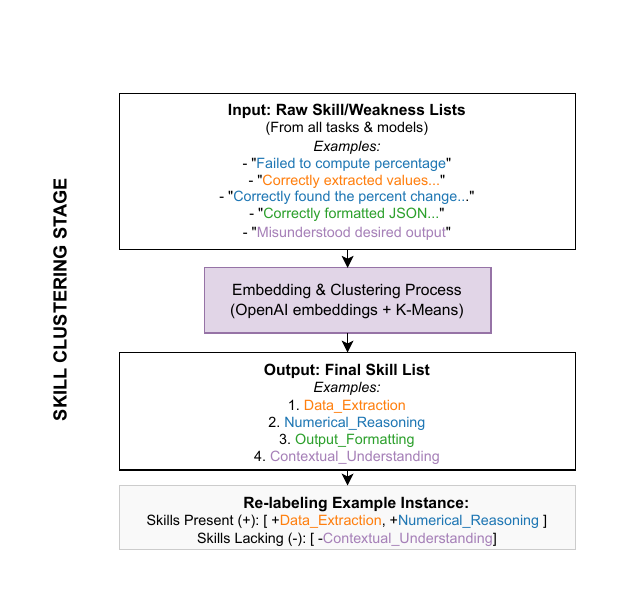}
\caption{Skill taxonomy construction. Raw skill descriptions are embedded, clustered, and assigned canonical labels to create a consistent vocabulary for capability representation.}
\label{fig:clustering}
\end{figure}

\bella~is a four-stage framework for skill-based model selection under deployment constraints. \bella's interpretable design decomposes selection process into modular, auditable components. \bella's four stages are: (1) benchmarking to assess cost and performance, (2) skill profiling via a critic LLM, (3) skill clustering, and (4) cost-aware model selection.

\vspace{-2 ex}
\subsection{Benchmarking}
We evaluate LLMs on multi-skill reasoning tasks, collecting: (1) \textbf{performance metrics} (accuracy, F1, etc); (2) \textbf{operational costs} (API pricing, latency); and (3) \textbf{model outputs} including reasoning traces to enable automated critique. This standardized evaluation enables fair comparison across heterogeneous models from small open-source alternatives to large proprietary closed-source models.

\subsection{Skill Profiling via Critic LLM}

For each model-task-instance triplet, a critic LLM extracts fine-grained skill profiles (Figure \ref{fig:profiling_example}). The critic receives the task input, reference solution, and model output, then produces a structured assessment identifying:

\begin{enumerate}[itemsep=1pt, topsep=3pt]
\item \textbf{Demonstrated skills:} Capabilities successfully exhibited (e.g., numerical calculation, temporal reasoning, fact verification).
\item \textbf{Missing skills:} Abilities the model failed to demonstrate which contributed to errors (e.g., incorrect data extraction leading to wrong input to calculation).
\item \textbf{Skill criticality:} How critical each demonstrated and missing skill was to the model's performance or lack thereof.
\end{enumerate}

This approach requires no predefined skill ontology and adapts to domain-specific capabilities, yielding rich instance-level annotations of model behavior. By grounding skill identification in actual model behavior rather than abstract taxonomies, we obtain annotations that reflect what models can actually do on specific instances.

\subsection{Skill Clustering}
The critic produces natural language skill descriptions (e.g., ``accurate arithmetic,'' ``mathematical computation,'' ``numerical precision''). To build consistent capability matrices, we canonicalize skills through:

\begin{enumerate}[itemsep=0pt, topsep=2pt,leftmargin=*]
\item \textbf{Embedding:} Encode skill phrases using a sentence transformer.
\item \textbf{Clustering:} Apply hierarchical clustering to group similar descriptions.
\item \textbf{Labeling:} Assign canonical labels via majority voting or LLM summarization.
\end{enumerate}

This yields a reduced taxonomy of interpretable skills. Each instance is re-encoded as a binary skill vector, enabling systematic comparison of model capabilities and task requirements.

\paragraph{Capability Matrices.}
The extraction and clustering stages produce three structured representations that encode the information needed for selection.

\begin{itemize}[nosep]
\item \textit{Model capability matrix $\mathbf{C} \in [0,1]^{M \times S}$.} Entry $C_{m,s}$ represents model $m$'s proficiency at skill $s$, computed as the fraction of skill-requiring instances where the model demonstrated that skill successfully.
\item \textit{Task requirement matrix $\mathbf{R} \in \{0,1\}^{T \times S}$.} Entry $R_{t,s}$ indicates whether task $t$ requires skill $s$, determined by aggregating skill annotations across instances.
\item \textit{Cost vector $\mathbf{c} \in \mathbb{R}^{M}$.} Entry $c_m$ represents the operational cost of model $m$ under the constraint of interest (monetary or latency).
\end{itemize}

\noindent
These matrices enable systematic reasoning about capability-requirement alignment under cost constraints.

\paragraph{Generalizability of the Matrix Framework.} \bella's capability matrices establish a flexible foundation that naturally supports multiple retrieval and recommendation paradigms.

\textit{(1) Similarity-based retrieval.} Given a task's skill requirements encoded as vector $\mathbf{r}_t \in \{0,1\}^S$, we can identify models with similar capability profiles through cosine similarity or Euclidean distance in the skill space. This approach treats model selection as a nearest-neighbor problem where $\text{sim}(m, t) = \frac{\mathbf{c}_m \cdot \mathbf{r}_t}{\|\mathbf{c}_m\| \|\mathbf{r}_t\|}$ ranks models by their alignment with required skills. Extensions to weighted similarity metrics can prioritize critical skills or account for skill interdependencies.

\textit{(2) Collaborative filtering.} The matrix structure parallels user-item rating matrices in recommender systems, where models correspond to users and skills to items. This analogy enables matrix factorization techniques to discover latent capability dimensions and predict model performance on unseen skill combinations. Non-negative matrix factorization $\mathbf{C} \approx \mathbf{U}\mathbf{V}^\top$ can reveal skills clusterings, allowing inference about model capabilities beyond direct observations.

\textit{(3) Performance estimation via supervised learning.} The capability-requirement matrices provide natural feature representations for training performance predictors. We can construct training examples where each pair $(m, t)$ is represented by skill-based features and labeled with observed performance. Multiple featurization strategies are possible: (1) inner product features $\langle \mathbf{r}_t, \mathbf{c}_m \rangle$ that capture alignment between requirements and capabilities; (2) concatenated features $[\mathbf{r}_t; \mathbf{c}_m]$ that allow learning complex interactions; (3) element-wise products $\mathbf{r}_t \odot \mathbf{c}_m$ that isolate relevant capabilities; or (4) higher-order polynomial features that model skill synergies. A classifier or regressor $f(\mathbf{r}_t, \mathbf{c}_m) \rightarrow \hat{p}(m, t)$ trained on historical evaluation data can then generalize to new task-model combinations, providing probabilistic performance estimates that inform the selection objective.

\textit{(4) Explainability through decomposition.} Unlike black-box selection heuristics, the matrix framework provides natural explanations for recommendations. When model $m$ is selected for task $t$, we can identify which skills $\{s : R_{t,s} = 1 \wedge C_{m,s} \geq \tau\}$ qualified the model, distinguished it from alternatives (via $\Delta C_{m,s} = C_{m,s} - \max_{m' \neq m} C_{m',s}$), and how cost-performance tradeoffs were resolved. This interpretability is crucial for building trust and enabling iterative refinement of skill taxonomies based on selection outcomes.

The generality of this representation extends beyond the specific selection algorithm: the same matrices support multi-objective optimization, portfolio selection across model ensembles, and dynamic updating as new evaluation data becomes available. By grounding model selection in explicit skill-level reasoning rather than opaque performance scores, the framework facilitates systematic analysis of model strengths, weaknesses, and complementarities.

\subsection{Skill-aware Model Selection}
Given a new task, \bella~performs cost-aware selection:
\begin{enumerate}[leftmargin=*]
    \item \textbf{Skill inference.} Required skills for $t$ are identified by either prompting a planner LLM with task examples or accepting user-specified requirements driven by domain expertise.

    \item \textbf{Capability filtering.} The model pool is filtered to those meeting proficiency thresholds on required skills:
    \[
    \mathcal{M}_\text{capable} = \{m : C_{m,s} \geq \tau \text{ for all } s \text{ where } R_{t,s} = 1\}
    \]
    
    \item \textbf{Cost-constrained selection.} Among capable models, we select one maximizing expected performance while respecting the budget:
    \begin{equation}
    \max_{m \in \mathcal{M}_\text{capable}} \quad \hat{p}(m, t) \quad \text{s.t.} \quad c_m \leq B
    \end{equation}
    where $\hat{p}(m, t)$ estimates performance based on capability profiles and task requirements, see Section 4.1.
    For multiple constraints (e.g., both latency and cost), we extend to Pareto optimization or weighted scalarization based on user priorities. The selection includes interpretable rationale: which skills the chosen model possesses, which constraints it satisfies, and how it compares to alternatives.
\end{enumerate}

\subsection{Evaluation Methodology}

We propose to evaluate \bella~using leave-one-out cross-validation on a corpus of financial benchmarks. In each fold, we synthesize capability matrices from $N-1$ datasets and test zero-shot generalization on the held-out task, inferring required skills directly from few-shot prompts. We compare \bella's budget-constrained recommendations against competitive baselines (RouteLLM, FrugalGPT) and an optimal oracle. Key metrics include total inference cost, accuracy gap relative to proprietary models, and selection precision. This process ensures robustness across diverse reasoning types, validated further by qualitative audits of the generated skill profiles to confirm accurate failure attribution.

\section{Discussion}

\bella~contributes a framework connecting interpretable skill profiling to cost-aware model selection. We discuss design considerations, limitations, and future research directions.

\subsection{Design Rationale}

The framework's central design choice is grounding selection in interpretable skills rather than learned embeddings or black-box routing policies. This choice involves trade-offs.

\paragraph{Interpretability versus optimization.} Skill-based selection provides transparent rationale---practitioners understand which capabilities drove a recommendation. Black-box routers may achieve tighter cost-performance trade-offs through end-to-end optimization, but at the cost of explainability. For deployment contexts requiring auditability or trust, interpretability may outweigh marginal performance gains.

\paragraph{Generalization versus specificity.} By extracting skills from model behavior rather than learning task-specific routing policies, \bella~aims to generalize to new tasks without retraining. The hypothesis, supported by Skill-Slices' finding that skill-based routing improves accuracy~\citep{moayeri2024skill}, is that skills transfer across tasks more reliably than learned policies.

\paragraph{Flexibility versus consistency.} Critic-based skill extraction adapts to domain-specific capabilities without predefined ontologies. This flexibility may introduce variance compared to structured taxonomies but enables application across diverse domains without manual feature engineering.

\subsection{Future Directions}

Several extensions could enhance \bella's capabilities:

\paragraph{Dynamic selection.} Complex tasks often comprise subtasks with varying skill requirements. Extending \bella~to select models dynamically during execution---using skill requirements that evolve as subtasks emerge---could capture additional value.

\paragraph{Domain adaptation.} Investigating skill transfer across domains (e.g., whether ``logical reasoning'' learned on legal tasks transfers to mathematical proofs) could enable few-shot adaptation with minimal profiling overhead.

\paragraph{Richer constraints.} Real deployments face multiple simultaneous constraints (cost, latency, energy, privacy). Developing multi-objective optimization with user-specified importance weights could better align recommendations with application priorities.

\subsection{Limitations}

We present \bella~as a framework design requiring empirical validation. Key areas of evaluation include measuring the performance of the critic LLM, including across different critic models, and investigating stability of the clustering-based skill taxonomy across different runs. The core hypothesis that skill-based selection outperforms alternatives under budget constraints requires comparison with RouteLLM, FrugalGPT, and oracle baselines. Skill profiling introduces upfront cost (critic calls for each model-task-instance). Characterizing this overhead and developing caching or incremental strategies would inform the practical deployment of \bella.

\section{Conclusion}

The proliferation of large language models creates both opportunity and challenge for practitioners seeking cost-effective deployment. Aggregate benchmarks provide high-level comparison but insufficient guidance for navigating cost-performance trade-offs under real-world constraints.

\bella~addresses this gap through interpretable skill profiling and constraint-aware model selection. The framework extracts fine-grained capability profiles from LLM outputs, organizes them into structured matrices, and formulates selection as constrained optimization over these representations. Unlike black-box routing systems, \bella~provides transparent rationale for recommendations---practitioners understand which capabilities a model possesses and how it satisfies their constraints.

By connecting capability assessment to deployment decisions, \bella~offers a principled approach to model selection that balances performance, cost, and interpretability. We hope this work contributes to making LLM deployment more accessible, efficient, and aligned with practitioner needs across diverse applications.

\subsection*{Acknowledgements}

This work was supported in part by the Financial Services and Innovation Lab at the Georgia Institute of Technology, led by Dr. Sudheer Chava.

\bibliography{bella_refs}
\bibliographystyle{unsrt}

\end{document}